\documentclass[letterpaper, 10 pt, conference]{ieeeconf}  
\IEEEoverridecommandlockouts                              

\usepackage{cite}
\usepackage{amsmath,amssymb,amsfonts}
\usepackage{algorithmic}
\usepackage{graphicx}
\usepackage{textcomp}
\def\BibTeX{{\rm B\kern-.05em{\sc i\kern-.025em b}\kern-.08em
    T\kern-.1667em\lower.7ex\hbox{E}\kern-.125emX}}
\usepackage{enumerate}
\usepackage{subcaption}
\usepackage[table,xcdraw]{xcolor}

\definecolor{PineGreen}{RGB}{0,153.0,0}

\title{\LARGE \bf
DOTIE - \underline{D}etecting \underline{O}bjects through \underline{T}emporal \underline{I}solation of \underline{E}vents using a Spiking Architecture
}

\author{Manish Nagaraj,
Chamika Mihiranga Liyanagedera and
Kaushik Roy
\thanks{*This work was supported in part by, Center for Brain-inspired Computing (C-BRIC), a DARPA sponsored JUMP center, Semiconductor Research Corporation (SRC), National Science Foundation, the DoD Vannevar Bush Fellowship, and IARPA MicroE4AI.}
\thanks{All authors are with Purdue University, West Lafayette, IN 47907, USA \{{\tt\small mnagara, cliyanag, kaushik}\}@purdue.edu} %
}

\begin{document}
\maketitle
\thispagestyle{empty}
\pagestyle{empty}

\begin{abstract}

Vision-based autonomous navigation systems rely on fast and accurate object detection algorithms to avoid obstacles. Algorithms and sensors designed for such systems need to be computationally efficient, due to the limited energy of the hardware used for deployment. Biologically inspired event cameras are a good candidate as a vision sensor for such systems due to their speed, energy efficiency, and robustness to varying lighting conditions. However, traditional computer vision algorithms fail to work on event-based outputs, as they lack photometric features such as light intensity and texture. In this work, we propose a novel technique that utilizes the temporal information inherently present in the events to efficiently detect moving objects. Our technique consists of a lightweight spiking neural architecture that is able to separate events based on the speed of the corresponding objects. These separated events are then further grouped spatially to determine object boundaries. This method of object detection is both asynchronous and robust to camera noise. In addition, it shows good performance in scenarios with events generated by static objects in the background, where existing event-based algorithms fail. We show that by utilizing our architecture, autonomous navigation systems can have minimal latency and energy overheads for performing object detection.

\end{abstract}
\section{Introduction}
\label{Sec:Intro}
Autonomous navigation is emerging as an important area of research, with applications ranging from videography and surveillance in remote regions to transportation systems in populated urban areas. The Society of Automation Engineers (SAE) has identified six levels of automation for autonomous navigation \cite{SAE:shadrin2019analytical}, starting from systems with no automation (\textit{level 0}) to systems that are fully automated and do not require any operator (\textit{level 5}). 
One of the most basic and essential requirements of a system  with automation (\textit{level 1} and above), is the ability to detect and avoid obstacles. And most importantly, these systems must be able to perform object detection accurately at very high speeds. 

Recent advancements in imaging technology have led to the development of biologically inspired \textit{event cameras} \cite{eventcamera1, eventcamera2, eventcamera3, eventcamera4}. While traditional \textit{frame cameras} capture photometric features such as light intensity and texture with high spatial resolution at fixed intervals, event cameras are asynchronous and only capture the change in light intensities at each pixel. Although event cameras fail at capturing photometric features, they do not suffer from issues such as motion blur and can operate at a much higher frequency and under a wider range of illumination. 
Owing to their higher output frequency, event cameras can capture high resolution temporal information that are missed by frame cameras. Since object detection and subsequent applications such as collision avoidance need to be performed at a very high speed and over a wide range of illumination, event cameras become ideal candidates for this task. 

\begin{figure}[htb]
\centering
\begin{subfigure}{0.3\columnwidth}
\centering
\includegraphics[width=\columnwidth]{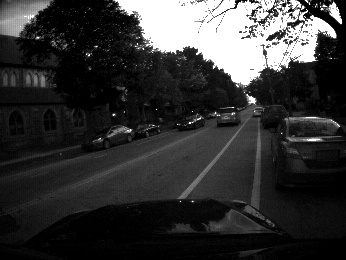}
\caption{\vspace{-1mm}}
\label{fig:grayimg_motiv}
\end{subfigure}
\begin{subfigure}{0.3\columnwidth}
\centering
\includegraphics[width=\columnwidth]{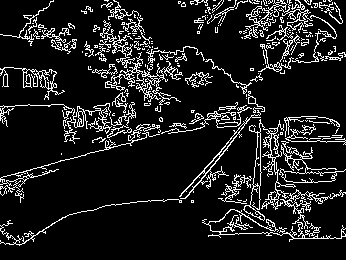}
\caption{\vspace{-1mm}}
\label{fig:gray_canny}
\end{subfigure}
\begin{subfigure}{0.3\columnwidth}
\centering
\includegraphics[width=\columnwidth]{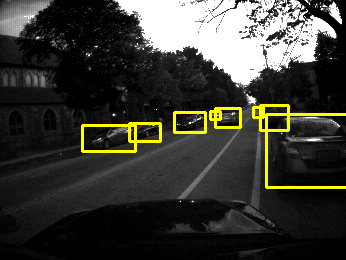}
\caption{\vspace{-1mm}}
\label{fig:gray_yolo_motiv}
\end{subfigure}\\[1ex]
\begin{subfigure}{0.3\columnwidth}
\centering
\includegraphics[width=\columnwidth]{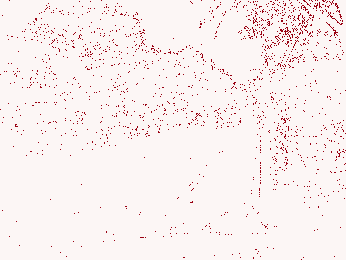}
\caption{\vspace{-1mm}}
\label{fig:event_motiv}
\end{subfigure}
\begin{subfigure}{0.3\columnwidth}
\centering
\includegraphics[width=\columnwidth]{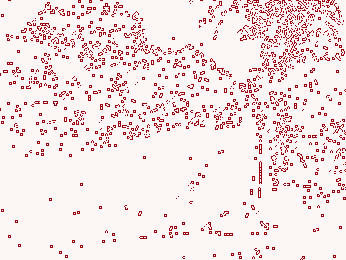}
\caption{\vspace{-1mm}}
\label{fig:event_canny}
\end{subfigure}
\begin{subfigure}{0.3\columnwidth}
\centering
\includegraphics[width=\columnwidth]{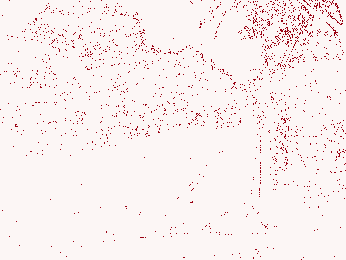}
\caption{\vspace{-1mm}}
\label{fig:event_yolo_moti}
\end{subfigure}
\caption{Example to show the inability of traditional computer vision algorithms to perform on event camera outputs. (a) and (d) show the frame and event camera outputs captured on the same scene (taken from the MVSEC dataset \cite{MVSEC:zihao2018multi}). (b) and (e) show the outputs of applying a canny filter on the respective images. (c) and (f) show the output of YOLOv3 on the respective images. We can see that both canny filtering and YOLOv3 fail to perform on the event camera output.}
\label{fig:canny_example}
\end{figure}

While there has been a plethora of research on object and feature detection algorithms, most of these algorithms are designed for frame camera outputs. These algorithms fail on event camera outputs as they rely on photometric features that are only present in frame data. For example, Fig. \ref{fig:canny_example}. shows the performance of both, a preliminary edge detection algorithm - the canny filter \cite{canny:1986computational}, and a complex learning based object recognition algorithm - YOLOv3 \cite{yolov3}, on a frame camera output and an event camera output of the same scene. While the algorithms can successfully function on the former, they fail to function on the latter. 


This motivates us to explore and utilize temporal features that are inherently present in the events, but are ignored in traditional computer vision algorithms. Leveraging such temporal information in events can help us in detecting and differentiating objects in a scene. For this purpose, we identify two properties important for object detection:
\begin{enumerate}[(i)]
    \item Events generated by the same object are temporally close to each other.
    \item Events generated by the same object are spatially close to each other.
\end{enumerate}

Based on these two properties, we aim to group events along two dimensions - spatial and temporal, to isolate events based on their source objects and thus identify the object boundaries. 


Biologically inspired spiking neurons, specifically \textit{Leaky Integrate and Fire} (LIF) neurons \cite{LIF:delorme1999}, can leverage spatio-temporal information and are hence, well suited for achieving boundary detection along both these dimensions. These neurons mimic the brain activity and behave based on the temporal properties of the inputs to the neuron. The neuron generates an output spike only if the input events occur at a rate higher than a certain frequency. A neural architecture with these spiking neurons is sensitive to the temporal structure of input events. It is also possible to arrange the connectivity between these neurons in a network to enable support between events that are spatially close to each other. 

Further, spiking neurons work in an asynchronous fashion, making them compatible with the asynchronous outputs of event cameras. These properties, along with the fact that spiking neural architectures can be more energy efficient (as shown in  \cite{spikeenergy1:han2020deep,spikeenergy2:kucik2021investigating, snnenergy:horowitz20141}) than their artificial neural counterparts, make them an ideal candidate for performing object detection with a low latency and energy overhead for autonomous navigation systems.

In this work, we develop a novel and energy efficient object detection technique to first isolate objects based on the speed of their movement. This is done using inputs from an event camera and a lightweight single layer spiking neural network. Once objects are separated based on their speed of movement, their corresponding events are then grouped together based on their spatial characteristics. For this, we utilize existing clustering techniques to further separate the events belonging to different objects, based on their spatial proximity. Here, we are taking advantage of the fact that the events from the same object, share similar temporal characteristics such as their speed of movement, and spatial characteristics such as being generated by pixels that are spatially close together. Our experiments show that this approach is a more efficient way to detect objects in terms of both latency and energy consumption.
By isolating events based on the speed of movement of their corresponding objects, we are also able to eliminate the non-relevant information caused by noise and background static objects\footnotemark. Further, this benefits the clustering techniques as they have a lower operational complexity due to the reduced number of samples (events) that need to be clustered. 
\footnotetext{If there is a need to detect static objects, this can be done by clustering the residual events that do not propagate through the spiking architecture.}

We summarize the main contributions as follows:
\begin{enumerate}
    \item We develop an object detection algorithm that solely relies on the event camera outputs and does not need any additional information from traditional frame cameras.
    \item Unlike many existing works on object detection which accumulate events for a time duration to create a frame, we perform object detection asynchronously as the events are being generated by the event camera.
    \item The proposed spike-based network for separating objects based on their motion consists of a single layer, resulting in a detection algorithm with lower latency and energy overhead.
    \item The outputs of the proposed spiking architecture (which isolates objects based on their speed), can be used with any spatial clustering technique that does not require prior knowledge of the number or the size of clusters.
    \item The spiking architecture is scene independent. This means that we do not have to train the parameters of the architecture based on the scene of deployment. These parameters directly correspond to the speed of the objects and can be fine-tuned prior to deployment.
\end{enumerate}

\section{Related Work}
\label{sec:Lit_survey}
\subsection{Object Detection in the Event Camera Domain}
Object detection is a topic that has been extensively studied in the computer vision community. There have been works ranging from simple feature detectors \cite{canny:1986computational, harris:1988combined}, to more complex learning-based methods \cite{hed:xie2015}. There has also been a significant interest in the learning community on neural networks that can not only detect objects, but can also classify them into different classes \cite{yolov3}. However, when it comes to autonomous navigation where object classification is not a priority, the latency and energy efficiency of the underlying algorithms take precedence. 

As discussed earlier, traditional frame-based algorithms fail to operate on event camera outputs due to the absence of photometric characteristics such as texture and light intensity. However, owing to the numerous advantages of event cameras, including higher operating speed, wider dynamic range, and lower power consumption, there has been a substantial interest in the community to develop algorithms that are more suited towards this domain.

Initial event-based detection algorithms such as \cite{earlyfeat1, earlyfeat2} were focused on detecting patterns present in the event camera output. The authors in \cite{blobdet}  used a simple blob detector to detect the inherent patterns present in the event data, and \cite{eventcorner} used a plane fitting method on the distribution of events to identify corners. A recent work adapted Gaussian mixture modelling to detect patterns in the event data \cite{GMM:pikatkowska2012spatiotemporal}. These methods, however, fail in scenarios where there are events generated by the background. As a solution, \cite{EED:mitrokhin2018event} proposed a motion compensation technique to eliminate events generated by the background, by estimating the system's ego-motion. The optimization involved, however, adds significant latency and computational overhead to the system.

To improve the detection accuracy, several recent research efforts were focused on utilizing information from both frame and event cameras \cite{grayandevents1, grayandevents2:eklt}. These hybrid methods detect features on the frames and track the objects through events. Since their detection relies on frame inputs, they cannot operate in scenarios with a wide dynamic range and are computationally expensive. 

There have also been efforts in creating dense descriptors or ``frames" by accumulating events for a pre-specified duration of time \cite{densedesc1:fast, densedesc2}. These frames can then be used with traditional computer vision algorithms. But these implementations show lower accuracy due to the incompatibility of the inputs. To overcome the problem, many learning approaches such as \cite{cnn1, cnn2, cnn3, cnn4, cnn5:YOLE} train Convolutional Neural Networks (CNNs) with such frames as inputs. However, such systems still add an overhead on the latency, and are not asynchronous in nature. Further, with the use of deep neural networks, there is an increased computational power overhead.

\subsection{Spiking Neural Architectures for Handling Events}
Spiking Neural Networks (SNNs) operate in an asynchronous fashion and can mitigate the latency issues experienced in learning-based techniques.
The authors in \cite{snnevent1} use an unsupervised learning technique to learn car trajectories on a freeway captured by an event camera. The works in \cite{snnevent2, snnevent3, snnevent4, hfirst, spikingyolo} utilize SNNs to not only detect objects, but also classify them. Such works, however, utilize deep SNNs that add a significant computational and latency overhead to the system. They also require intensive training techniques and lose accuracy when they are deployed in a scene different from the one used to train the network.

In this paper, we propose a network with only a single layer of spiking neurons to capture the temporal information present in event data. This mitigates the energy and latency overhead of the neural architecture. Further, this network does not need to be trained on an extensive amount of data, but rather the hyperparameters are simply fine-tuned prior to deployment. 
Unlike other works, we observe that there is no accuracy degradation when our algorithm is tested on different scenes of deployment.

\subsection{Spatial Clustering of Events}
Clustering techniques such as \cite{kmeans:hartigan1979algorithm, DBSC:duan2007local, meanshift_orig} prove useful in grouping samples with a low dimensional complexity. Since events can be considered as samples that have spatial and temporal dimensions, there have been attempts to utilize clustering techniques for grouping events and detecting the corresponding object boundaries. \cite{meanshift:chen2018} uses Mean-shift clustering, proposed in \cite{meanshift_orig}, to cluster events, while \cite{GSCE:mondal2021moving} uses a graph spectral clustering technique to group events to determine object boundaries. However, such techniques assume that only the moving objects generate events. In the method proposed here, we first eliminate the events generated from the background and camera noise enabling the use of spatial clustering techniques. Since both the number and the size of the objects are unknown before deployment, we employ clustering techniques that are independent of such parameters \cite{DBSC:duan2007local}.

\section{Methodology}

\subsection{Separating Objects in the Temporal Domain}
As mentioned in Section \ref{Sec:Intro}, in order to detect objects efficiently, we aim to isolate objects along both the temporal and spatial domains. Spiking neurons are an excellent candidate as they are capable of operating along the temporal domain. The LIF neuron model \cite{LIF:delorme1999} has two characteristic features. One is an internalized state called \textit{membrane potential} ($U[t]$) and the other is a decay factor known as \textit{leak factor} ($\beta$). The neuron is initialized with a \textit{threshold value}, $U_{thr}$, and an initial membrane potential, $U[t_0]$. The weighted sum of the inputs ($WX[t]$) at each time step $t$, is accumulated in the membrane potential of the neuron as shown by Eqn~(\ref{eqn:spiking}).  The accumulated membrane potential decays over time and the decay rate is controlled by the leak factor. The leak factor denotes how much of the membrane potential is retained for the next time step, i.e., the higher the leak factor, the slower the rate of decay. If the accumulated membrane potential of the neuron exceeds the threshold at any point, ($U[t_n]>U_{thr}$), the neuron emits an output spike and resets its membrane potential.
\begin{equation}
    U[t_n] = \underbrace{\beta U[t_{n-1}]}_\text{decaying membrane potential}  + \underbrace{WX[t]}_\text{weighted sum of inputs}
    \label{eqn:spiking}
\end{equation}

\begin{figure}[htb]
    \centering
    \includegraphics[width=\columnwidth]{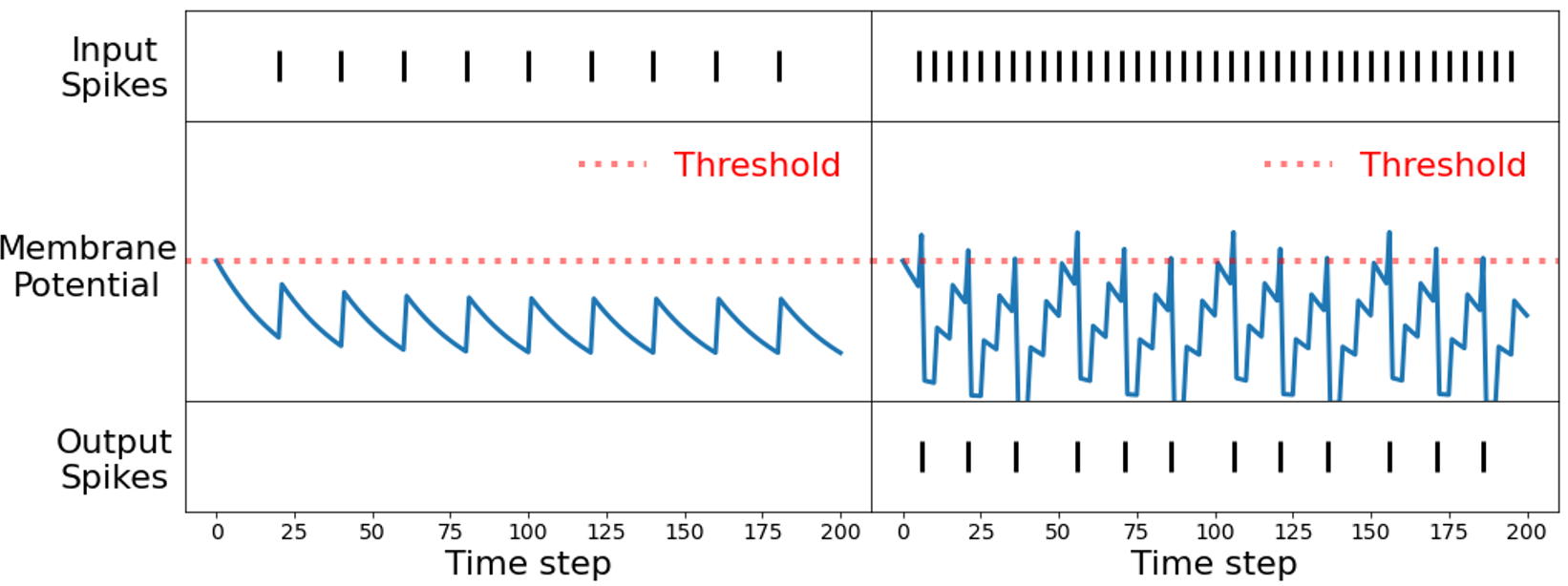}
    \caption{Temporal sensitivity of a spiking neuron. The graphs on the left shows the neuron response to inputs occurring at a lower rate, the graphs on the right shows the same neuron respond to inputs occurring at a higher rate. Once the inputs occur at a rate higher than a certain frequency, the neuron is able to generate outputs.}
    \label{fig:Spike_speed_demo}
    \vspace{-4mm}
\end{figure}

Such neurons are sensitive to the temporal information present in the input data. By tuning the hyperparameters (threshold and leak factor), it is possible to identify input spikes that are close to each other in time. This phenomenon is demonstrated in Fig. \ref{fig:Spike_speed_demo}. If the time between input spikes is large, the membrane potential decays its value before it can reach the threshold. However, if these inputs occur more frequently, they are able to overcome the decay and increase the membrane potential towards the threshold. Thus, the neuron generates output spikes if the input events occur at a frequency higher than a certain value. On the other hand, event cameras generate events when the light intensity acting on the pixels change with time. The rate at which these events are generated will depend on how fast the scene or objects are moving. Hence, a faster moving object will be responsible for generating events at a much higher frequency. When these properties of spiking neurons and event data are combined together, we have a network that is sensitive to objects that are moving faster than a certain speed.

\begin{figure}[htb]
\centering
\begin{subfigure}{\columnwidth}
\centering
\includegraphics[width=\columnwidth]{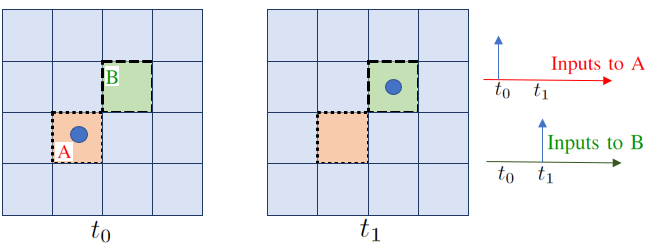}
\caption{Each neuron is connected to a camera pixel directly.}
\label{fig:neighbscena}
\end{subfigure}\\[1ex]
\begin{subfigure}{\columnwidth}
\centering
\includegraphics[width=\columnwidth]{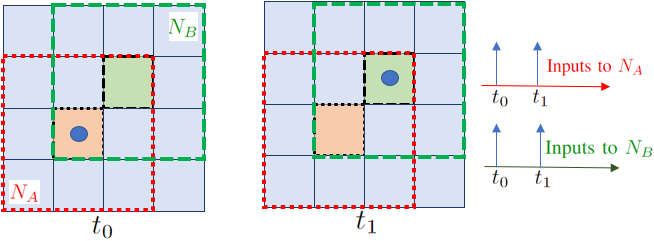}
\caption{Each neuron is connected to a neighborhood of pixels.}
\label{fig:neighbscenb}
\end{subfigure}
\caption{Two scenarios, with events generated identically at two consecutive time steps. In scenario (a), the events are registered as inputs only once at each neuron. In scenario (b), the events are registered in both time steps at both neurons.}
\label{fig:neighb_expl}
\vspace{-2mm}
\end{figure}

While directly connecting each pixel of the event camera to a spiking neuron will measure the frequency of inputs at that pixel, it does not necessarily measure the speed of an object. Let us consider the scenarios shown in Fig. \ref{fig:neighb_expl}. Here, an object generates an event at pixel A at time~step~$t_0$. If the object is moving at a fast speed, the event is generated in the neighboring pixel B at the next time~step~$t_1$. If we connect each pixel to a single spiking neuron, then both the neurons connected to pixels A and B only receive one input at each time step (Fig. \ref{fig:neighbscena}). Thus, the neural architecture cannot identify the fast-moving object. However, if we use a weighted sum from a neighborhood of pixels as inputs to neurons (Fig. \ref{fig:neighbscenb}), then the neurons connected to neighborhoods $N_A$ and $N_B$ (centered around pixels $A$ and $B$ respectively) register two consecutive inputs. This leads to the spiking architecture successfully recognizing the object as fast moving. However, using a very large neighborhood would lead to events corresponding to different objects that are spatially apart, contribute to the same neuron, preventing object isolation. Hence, it is essential to choose an appropriate neighborhood size. We found that a neighborhood of $(3\times 3)$ produced the best results. 

\begin{figure}[htb]
    \centering
    \includegraphics[width=\columnwidth]{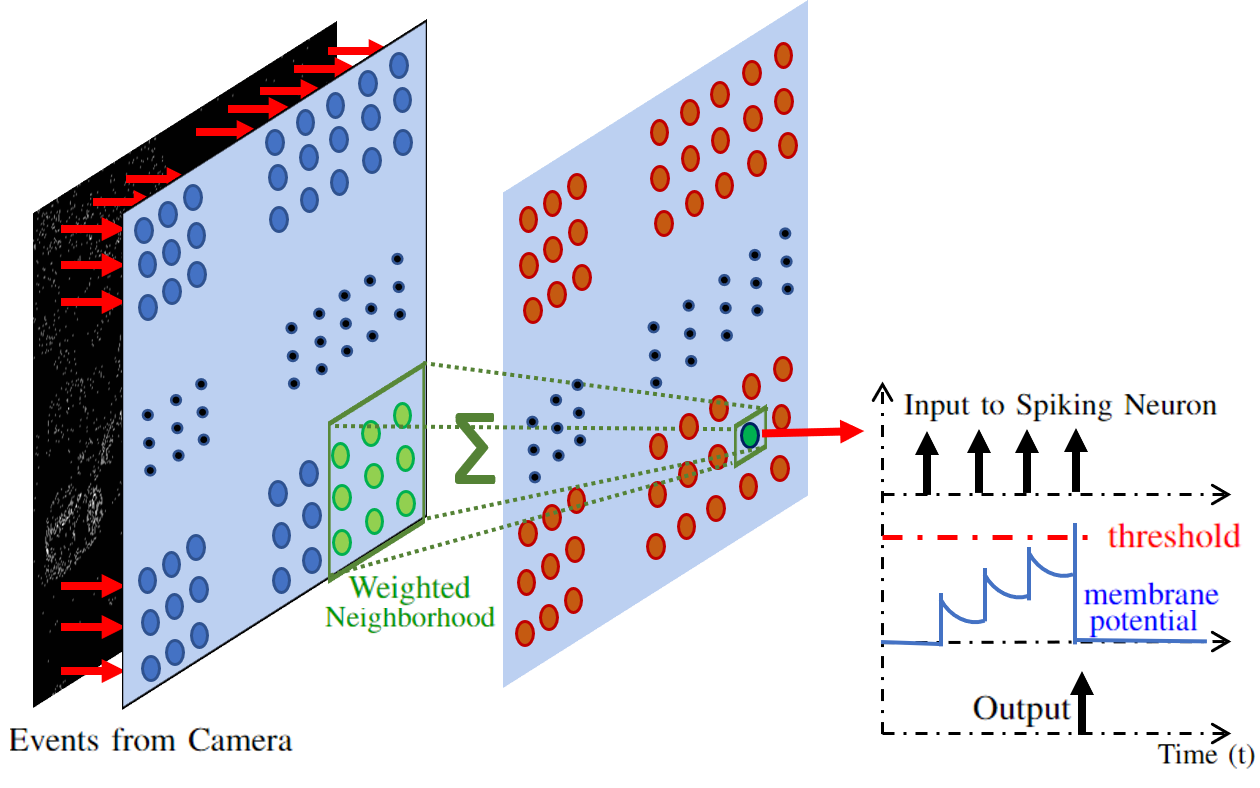}
    \caption{Spiking architecture used to detect and isolate events belonging to objects moving at a faster speed. A weighted summation of inputs from pixels of the event camera are directly fed into the neurons. If the object is moving at a higher speed, it will prompt the neuron to generate a spiking output. This can be used as a detecting gate, that only propagates events from objects moving at a higher speed and filter out the remaining events.}
    \label{fig:spike_arch}
    \vspace{-5mm}
\end{figure}

Therefore, as demonstrated in Fig. \ref{fig:spike_arch}, we consider a $(3 \times 3)$ weighted neighborhood of pixels as inputs to each neuron. The weights for the neighborhood were normalized to sum to $1$. The central pixel was assigned a weight of $0.2$, while the rest of the pixels were assigned equal weights of $0.1$. This allowed our architecture to detect higher speed objects in a scene more efficiently.

\begin{figure}[htb]
\centering
\begin{subfigure}{.45\linewidth}
\centering
\includegraphics[width=\columnwidth]{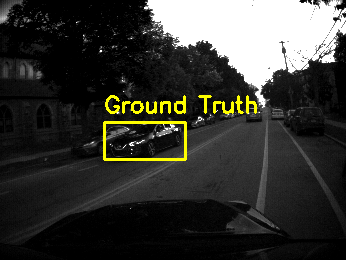}
\caption{Input grayscale image}
\label{fig:steps_gray}
\end{subfigure}
\begin{subfigure}{.45\linewidth}
\centering
\includegraphics[width=\columnwidth]{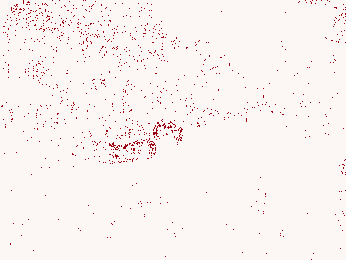}
\caption{Input events}
\label{fig:steps_events}
\end{subfigure} \\[1ex]
\begin{subfigure}{.45\linewidth}
\centering
\includegraphics[width=\columnwidth]{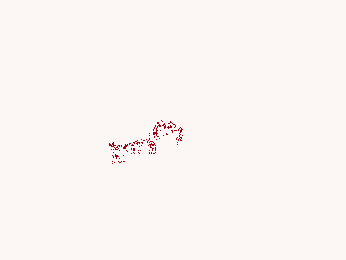}
\caption{Isolated moving object}
\label{fig:steps_isolated}
\end{subfigure}
\begin{subfigure}{.45\linewidth}
\centering
\includegraphics[width=\columnwidth]{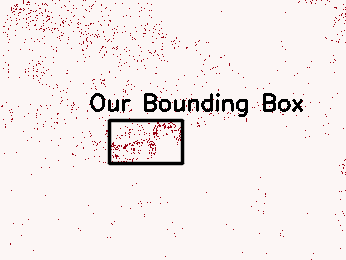}
\caption{Resulting bounding box}
\label{fig:steps_complete}
\end{subfigure}
\caption{Demonstration of our algorithm on the MVSEC dataset. The events corresponding to the moving object are isolated and then grouped together based on spatial proximity.}
\label{fig:separation_demo}
\vspace{-6mm}
\end{figure}

Once we use the neural architecture to detect fast moving objects, we then need to isolate the input events that correspond to these objects before we separate them in the spatial domain. In order to do this, we recover the inputs around neighborhoods of the spike outputs. The size of this recovery neighborhood can be tuned and is not restricted to the size of the input neighborhood for neurons. Our spiking layer will act asynchronously with a delay of one time step (to check for the output spike and recover the inputs in the previous time step that caused this output). This method, unlike many existing works, does not require event data to be accumulated over a period of time before they can be used for processing. Fig. \ref{fig:separation_demo}. demonstrates the operation of the proposed spiking architecture using a scene from the MVSEC dataset \cite{MVSEC:zihao2018multi}.



\subsection{Clustering in the Spatial Domain}
In order to detect and draw accurate object boundaries, we use spatial clustering on the separated input events. Although there are many clustering techniques that are widely available, most of them require us to either know the size of the clusters or the number of clusters in the data prior to clustering. Since we do not have any prior knowledge of the input scenario, we need a clustering technique that is independent of both these factors. For our experiments, we chose a density-based technique called Density Based Spatial Clustering (DBSC) \cite{DBSC:duan2007local}. The only parameters this technique requires are the minimum number of data points needed to form a cluster and the minimum data density of a cluster. Both of these parameters were fine tuned for the best results. 

\subsection{Extending to  Multiple Speeds}
\begin{figure}[htb]
\vspace{-3mm}
    \centering
    \includegraphics[width=\columnwidth]{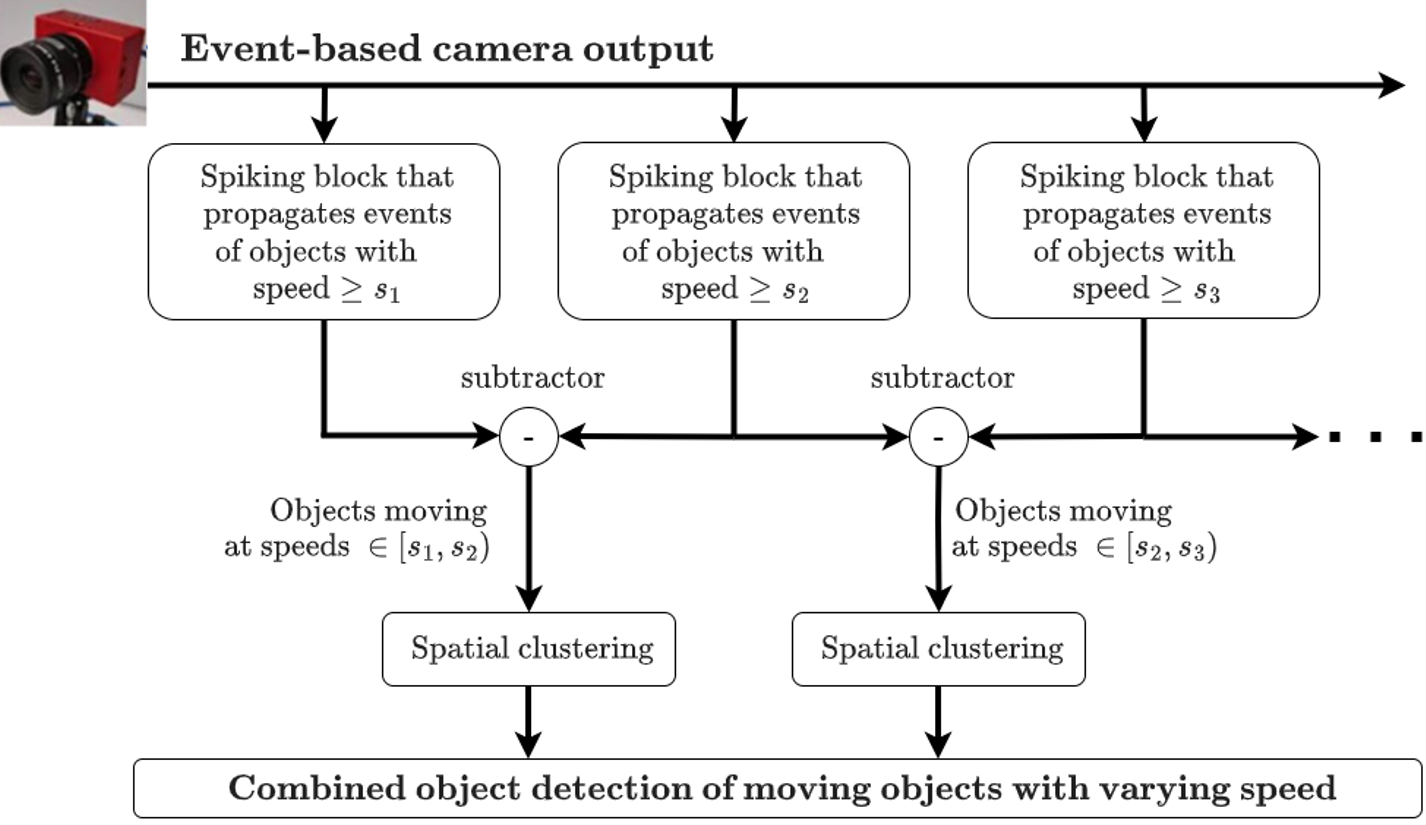}
    \caption{Overview of the architecture to detect multiple objects moving at varying speeds. Each of the spiking blocks is fine-tuned to a specific range of speeds. Events corresponding to objects moving at a speed within this range are propagated. These input events can then be spatially clustered in order to detect object boundaries.}
    \label{fig:overview_arch}
    \vspace{-3mm}
\end{figure}

So far, we have proposed a spiking neural block that can separate out objects moving from those that are stationary. While this is useful and already shows improvements at object detection in real world scenarios, we can further separate objects based on their speed. Fig. \ref{fig:overview_arch}. shows the overview of the architecture that can be used to separate objects based on their speeds.

Each branch of this architecture will isolate objects moving faster than a  certain speed. By considering the difference between the outputs of branches that have successive threshold speeds $s_1$ and $s_2$, we can isolate objects that are moving with a speed in the range $[s_1, s_2)$. Multiple such ranges could be added and fine-tuned before deployment, based on the application. These branches operate in parallel and will not add any latency. Since each branch only consists of a single layer of spiking neurons, the overall energy efficiency of the system is still preserved.

\section{Experiments and Results}
\label{Sec:Exp}
In this section, we both qualitatively and quantitatively show the accuracy and energy efficiency of our proposed technique. Most existing works evaluate their algorithms on private datasets that are generated by the authors. These datasets are not available to the public and do not resemble event camera outputs that are generated in real life scenarios. In order to measure the accuracy and energy efficiency of our algorithm in real scenarios, we chose the MVSEC dataset \cite{MVSEC:zihao2018multi} to show our results. This is a publicly available dataset and contains data collected by both an event camera, and a traditional frame camera mounted on a car.

\subsection{Accuracy}
As stated earlier, most existing works can only operate in scenarios without any background or camera noise. Our algorithm however can work well under these conditions. This is because, the spiking architecture acts as a temporal filter and can eliminate both noisy inputs from the camera as well as the events generated from the static objects in the background. 

Since the MVSEC dataset does not contain ground truth bounding boxes of objects in the foreground, we used YOLOv3 \cite{yolov3}, a traditional frame-based state of the art algorithm to generate bounding boxes on the corresponding grayscale image. We used this as the ground truth and then compared the \textit{intersection over union} ($IoU$) of the objects present in the foreground (not static). Since the grayscale images are generated at a much lower rate, we only calculate $IoU$s of the event-based detection algorithms at these time frames. It is to be noted, however, that our detection algorithm is asynchronous and can operate at the rate events are generated by the event camera. 

We consider a bounding box with an $IoU \geq 0.5$ with respect to the ground truth as a \textit{true positive} ($TP$) and one with an $IoU < 0.5$ as a \textit{false negative} ($FN$). If a bounding box was generated when no corresponding ground truth was detected, we considered this a \textit{false positive} ($FP$). The \textit{precision} and \textit{recall} 
were calculated based on their standard definitions using $TP$, $FN$, and $FP$. Table \ref{Tab:acc} shows the comparison of these accuracy metrics of detection algorithms on the \textit{outdoor day 2} segment of the MVSEC dataset. 

\begin{table}[htbp]
\caption{Table showing the comparison of accuracy metrics of existing object detection algorithms to our algorithm. Experiments were conducted on the \textit{outdoor day 2} segment of the MVSEC dataset.}
\vspace{-2mm}
 \renewcommand*{\arraystretch}{1.2}
\begin{center}
\begin{tabular}{|c|c|c|c|}
\hline
\textbf{Algorithm} & \multicolumn{1}{c|}{\textbf{Mean $IoU$}} & \multicolumn{1}{c|}{\textbf{Recall}} & \textbf{Precision} \\ \hline \hline
GMM \cite{GMM:pikatkowska2012spatiotemporal} & $0.2154$ & $0.08$ & $1.00$  \\ \hline
Meanshift \cite{meanshift:chen2018}         & $0.3986$ & $0.23$ & $1.00$  \\ \hline
K-Means \cite{kmeans:hartigan1979algorithm} & $0.0997$ & $0.00$ & $0.00$  \\ \hline
GSCE  \cite{GSCE:mondal2021moving}          & $0.1071$ & $0.00$ & $0.00$  \\ \hline
DBSCAN only \cite{DBSC:duan2007local}       & $0.1244$ & $0.00$ & $0.00$  \\ \hline
\rowcolor[HTML]{C0C0C0} 
\textbf{Ours (+ DBSCAN)}                             & $\textbf{0.8593}$ & $\textbf{1.00}$ & $\textbf{1.00}$  \\ \hline
\end{tabular}
\label{Tab:acc}
\end{center}
\vspace{-3mm}
\end{table}

\begin{figure}[!htb]
\vspace{-2.5mm}
\centering
\begin{subfigure}{.45\linewidth}
\centering
\includegraphics[width=\columnwidth]{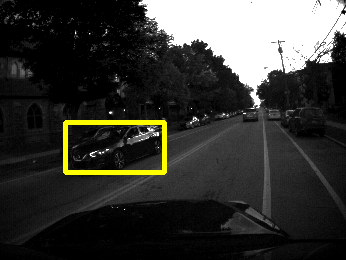}
\caption{Grayscale frame}
\label{fig:comp_gray}
\end{subfigure}
\begin{subfigure}{.45\linewidth}
\centering
\includegraphics[width=\columnwidth]{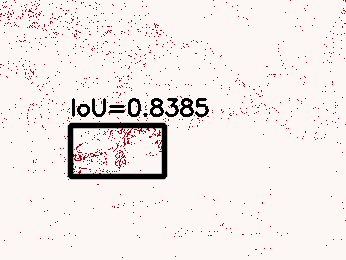}
\caption{Our algorithm}
\label{fig:comp_ours}
\end{subfigure}\\[1ex]
\begin{subfigure}{.45\linewidth}
\centering
\includegraphics[width=\columnwidth]{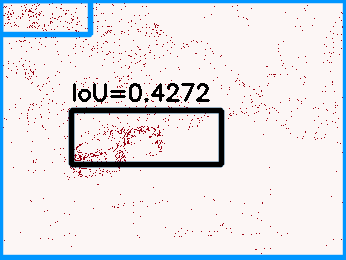}
\caption{Meanshift \cite{meanshift:chen2018}}
\label{fig:comp_meanshift}
\end{subfigure}
\begin{subfigure}{.45\linewidth}
\centering
\includegraphics[width=\columnwidth]{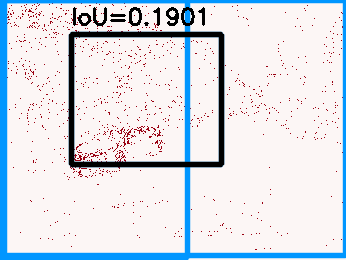}
\caption{GMM \cite{GMM:pikatkowska2012spatiotemporal}}
\label{fig:comp_gmm}
\end{subfigure}\\[1ex]
\begin{subfigure}{.45\linewidth}
\centering
\includegraphics[width=\columnwidth]{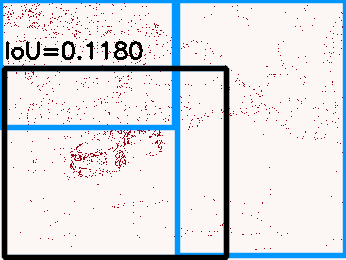}
\caption{K-Means \cite{kmeans:hartigan1979algorithm}}
\label{fig:comp_kmeans}
\end{subfigure}
\begin{subfigure}{.45\linewidth}
\centering
\includegraphics[width=\columnwidth]{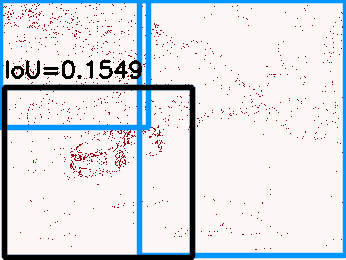}
\caption{GSCE \cite{GSCE:mondal2021moving}}
\label{fig:comp_gsce}
\end{subfigure}\\[1ex]
\begin{subfigure}{\linewidth}
\centering
\includegraphics[width=0.45\columnwidth]{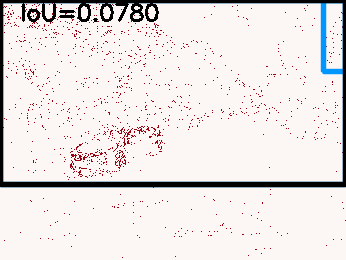}
\caption{DBSCAN only \cite{DBSC:duan2007local}}
\label{fig:comp_dbsc}
\end{subfigure}
\caption{Comparison of our object detection algorithm to existing algorithms. In case of algorithms that generate multiple bounding boxes, the box with the highest $IoU$ with respect to the ground truth is considered. These boxes are highlighted in black, and the other bounding boxes are highlighted in blue\protect\footnotemark. }
\label{fig:comp_all}
\vspace{-6mm}
\end{figure}

Fig. \ref{fig:comp_all}. shows a more qualitative comparison of the bounding boxes predicted by each of these algorithms. We observe that utilizing clustering techniques (\hspace{-0.01mm}\cite{DBSC:duan2007local, kmeans:hartigan1979algorithm}) directly on the event camera output fails to detect objects accurately. This is because, the events generated by the background and camera noise can occur spatially close to the events generated by the foreground (or moving objects).  Other techniques such as \cite{meanshift:chen2018} and \cite{GMM:pikatkowska2012spatiotemporal} are able to minimize the number of false positives, but perform poorly producing a large number of false negatives (boxes with $IoU < 0.5$). These  algorithms also generate multiple bounding boxes even if there is only one object in the foreground.  Our algorithm outperforms all the other existing algorithms at all instances of the dataset, and minimizes the number of false positives and false negatives while producing bounding boxes with a good $IoU$ ($\geq 0.5$). For a fairer comparison, we consider the bounding box with the highest $IoU$ with respect to the ground truth for computing the accuracy metrics in algorithms that generate multiple bounding boxes.

\footnotetext{Note that when the algorithms fail to isolate the moving object (7c)-(7g), there could be bounding boxes that cover the entire frame.}

\subsection{Computational Efficiency}
One of the main benefits of the proposed approach of object detection is the computational efficiency in terms of both energy and latency. Spiking neurons accumulate inputs in the membrane potential and only emit an output spike if the accumulated membrane potential exceeds a set threshold. These \textit{accumulate} (AC) operations require only a fraction of the energy needed for computation when compared to the traditional \textit{multiply and accumulate} (MAC) operations that are used in artificial neural networks. It has been shown in \cite{snnenergy:horowitz20141}, that for $32$-bit floating-point computations performed on a $45nm$ CMOS technology, the energy for AC operations ($E_{AC}$) is $0.9pJ$ and the energy for MAC operations ($E_{MAC}$) is $4.6pJ$. Further, due to the nature of events, these AC operations occur in a sparse fashion. 

We can estimate the energy consumed for computation by the spiking layer by multiplying the number of such synaptic operations \#OPS\textsubscript{SNN} with $E_{AC}$. Formally, the total computational energy is given as:
\begin{equation}
    \text{E\textsubscript{spiking}} = \text{\#OPS\textsubscript{SNN}} \times E_{AC} =
    M\times C\times F \times E_{AC}
\end{equation}
 where $M$ is the number of spiking neurons, $C$ is the number of synaptic connections (inputs to each neuron) and $F$ is the mean pre-spiking input rate. By plugging in the appropriate values ($M=260\times346, C=3\times3, F=0.0151$), we estimate our architecture consumes an average energy of $11.03nJ$, during inference on the MVSEC \textit{outdoor day 2} segment.
 This is a huge reduction in energy consumption compared to average energy of $44.06mJ$ that a standard artificial neural network YOLOv3 \cite{yolov3} uses during inference on the same dataset. We estimated the latter value by multiplying the number of MAC operations in the entire network ($\sum_{layer}M_{layer}\times C_{layer} = 9.58\times10^9$) with the energy consumption of a single MAC operation. 



Our proposed technique is also efficient in terms of latency. While traditional computer vision works such as \cite{yolov3} are limited by the input rate of frame cameras, our algorithm can operate at much faster rates owing to the high output rate of event cameras. We also do not need to accumulate events for a duration of time before processing, unlike many event-based detection algorithms \cite{EED:mitrokhin2018event, densedesc1:fast, densedesc2}. This true asynchronous nature of the algorithm minimizes our processing overhead in terms of latency.


\section{Conclusion}
In this work, we present a novel object detection algorithm for autonomous navigation tasks using biologically inspired spiking neurons in combination with an event camera. By using the inherent temporal information present in the input data, the proposed spiking architecture isolates events based on the speed of objects. 
We show our algorithm excels at detecting object boundaries, even in scenarios with camera noise and background objects, where other existing event-based algorithms fail. The proposed methodology carries low energy overhead, owing to the sparsity of input data, low network complexity and the computationally efficient accumulate (as opposed to multiply and accumulate in standard deep learning)
operations of spiking neurons. In addition, the spiking architecture developed is independent of the scene of deployment, and is scalable to multiple speeds.



\newpage
\bibliographystyle{IEEEtran}
\bibliography{References}
\end{document}